\documentclass[10pt,twocolumn,letterpaper]{article}

\usepackage{cvpr}
\usepackage{times}
\usepackage{epsfig}
\usepackage{graphicx}
\usepackage{amsmath}
\usepackage{amssymb}

\usepackage[table]{xcolor}
\usepackage{color}
\usepackage{rotating}
\usepackage{array}

\usepackage{tikz}
\usetikzlibrary{spy, shapes, arrows, matrix, pgfplots.groupplots}
\usetikzlibrary{decorations.text}
\definecolor{myblue}{rgb}{.1,0.3,0.9}

\definecolor{myorchid}{RGB}{200,10,10}
\definecolor{myblue}{RGB}{10,50,200}
\definecolor{darkgreen}{RGB}{10,150,10}
\definecolor{darkred}{RGB}{200,50,20}
\definecolor{YellowOrange}{RGB}{255,127,39}

\usepackage{booktabs}
\usepackage{xfrac}
\usepackage{tablefootnote}
\usepackage{colortbl}
\usepackage{tabularx}
\definecolor{rowblue}{RGB}{220,230,240}

\usepackage{gensymb}

\newcommand{\comment}[1]{}

\renewcommand{\paragraph}[1]{\vspace{6px} \noindent \textbf{#1} \ \ }

\definecolor{chblue}{RGB}{40,121,215}
\definecolor{chgray}{RGB}{100,100,100}

\usepackage{amsfonts}
\usepackage{color}
\usepackage{caption}
\usepackage{comment}
\usepackage{graphicx}
\usepackage{amsmath}
\usepackage{amssymb}
\usepackage{mathrsfs}
\usepackage{multirow}
\usepackage{floatrow}

\newcommand{\I}{\mathcal{I}}

\newcommand{\M}{\mathcal{M}}
\newcommand{\p}{\mathbf{p}}
\newcommand{\0}{\mathbf{0}}
\newcommand{\deltap}{\Delta \p}
\newcommand{\Ifg}{\I_\text{FG}}
\newcommand{\Ibg}{\I_\text{BG}}
\newcommand{\Mfg}{\M_\text{FG}}
\newcommand{\Icomp}{\I_\text{comp}}
\newcommand{\Ireal}{\I_\text{real}}

\newcommand{\G}{\mathcal{G}}
\newcommand{\D}{\mathcal{D}}

\newcommand{\E}{\mathbb{E}}
\renewcommand{\H}{\mathbf{H}}
\newcommand{\loss}{\mathcal{L}}

\newcommand{\Pset}{\mathbb{P}}
\newcommand{\X}{\mathbf{X}}

\usepackage[pagebackref=true,breaklinks=true,letterpaper=true,colorlinks,bookmarks=false]{hyperref}

\cvprfinalcopy 

\def\cvprPaperID{380} 

\ifcvprfinal\fi
\begin{document}

\title{ST-GAN: Spatial Transformer Generative Adversarial Networks \\ for Image Compositing}

\author{
Chen-Hsuan Lin\textsuperscript{1\ref{fn}} \quad
Ersin Yumer\textsuperscript{2,3\ref{fn}} \quad
Oliver Wang\textsuperscript{2} \quad
Eli Shechtman\textsuperscript{2} \quad
Simon Lucey\textsuperscript{1,3} \\
\textsuperscript{1}Carnegie Mellon University \quad
\textsuperscript{2}Adobe Research \quad
\textsuperscript{3}Argo AI \\
{\tt\small chlin@cmu.edu \quad meyumer@gmail.com \quad \{owang,elishe\}@adobe.com \quad slucey@cs.cmu.edu}
}

\maketitle
\renewcommand*{\thefootnote}{\fnsymbol{footnote}}
\setcounter{footnote}{1}
\footnotetext{Work done during CHL's internship at Adobe Research.\label{fn}}
\renewcommand*{\thefootnote}{\arabic{footnote}}
\setcounter{footnote}{0}


\begin{abstract}

We address the problem of finding realistic \emph{geometric corrections} to a foreground object such that it appears natural when composited into a background image.
To achieve this, we propose a novel Generative Adversarial Network (GAN) architecture that utilizes Spatial Transformer Networks (STNs) as the generator, which we call Spatial Transformer GANs (ST-GANs). 
ST-GANs seek image realism by operating in the geometric warp parameter space. 
In particular, we exploit an iterative STN warping scheme and propose a sequential training strategy that achieves better results compared to naive training of a single generator. 
One of the key advantages of ST-GAN is its applicability to high-resolution images indirectly since the predicted \emph{warp parameters} are transferable between reference frames. 
We demonstrate our approach in two applications: (1) visualizing how indoor furniture (\eg from product images) might be perceived in a room, (2) hallucinating how accessories like glasses would look when matched with real portraits. 

\end{abstract}


\section{Introduction}

\begin{figure}[t!] \center
\includegraphics[width=0.97\linewidth]{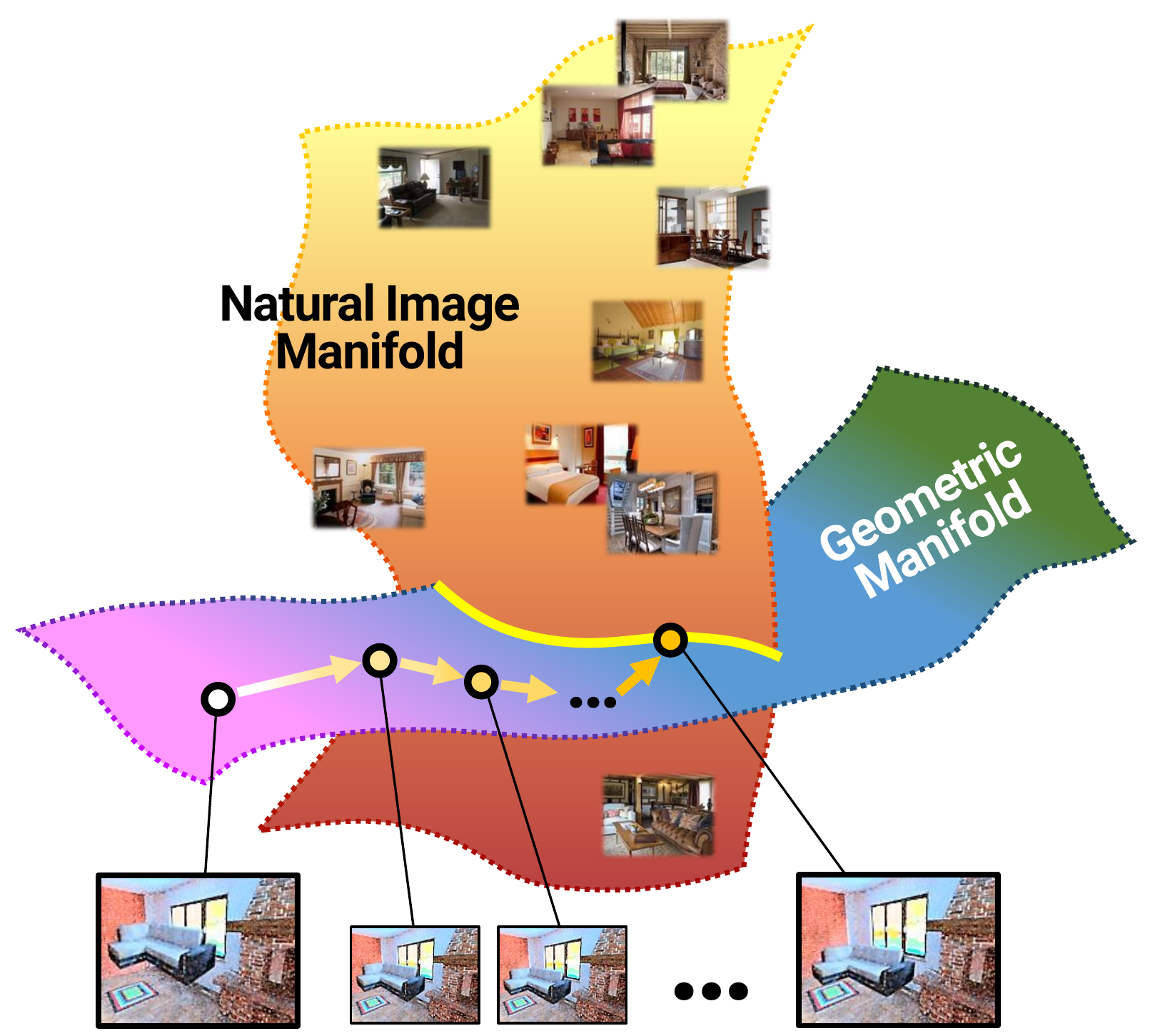}
\caption{Composite images easily fall outside the natural image manifold due to appearance and geometric discrepancies.
We seek to learn \textit{geometric corrections} that sequentially warp composite images towards the intersection of the geometric and natural image manifolds.
}
\label{fig:teaser}
\end{figure}

Generative image modeling has progressed remarkably with the advent of convolutional neural networks (CNNs).
Most approaches constrain the possible appearance variations within an image by learning a low-dimensional embedding as an encoding of the natural image subspace and making predictions from this at the pixel level.
We refer to these approaches here as \textit{direct image generation}.
Generative Adversarial Networks (GANs)~\cite{goodfellow2014generative}, in particular, have demonstrated to be an especially powerful tool for realistic image generation.
They consist of a generator network ($\G$) that produces images from codes, and a discriminator network ($\D$) that distinguishes real images from fake ones.
These two networks play a minimax game that results in $\G$ generating realistic looking images and $\D$ being unable to distinguish between the two when equilibrium is reached.

Direct image generation, however, has its limitations.
As the space of all images is very high-dimensional and image generation methods are limited by finite network capacity, direct image generation methods currently work well only on restricted domains (\eg faces) or at low resolutions.

In this work, we leverage Spatial Transformer Networks (STNs)~\cite{jaderberg2015spatial}, a special type of CNNs capable of performing geometric transformations on images, to provide a simpler way to generate realistic looking images -- by restricting the space of possible outputs to a well-defined \emph{low-dimensional geometric transformation} of real images.
We propose Spatial Transformer Generative Adversarial Networks (ST-GANs), which learn Spatial Transformer generators within a GAN framework.
The adversarial loss enables us to learn geometric corrections resulting in a warped image that lies at the \textit{intersection} of the natural image manifold and the geometric manifold -- the space of geometric manipulations specific to the target image (Fig.~\ref{fig:teaser}).
To achieve this, we advocate a sequential adversarial training strategy to learn iterative spatial transformations that serve to break large transformations down into smaller ones. 

We evaluate ST-GANs in the context image compositing, where a source foreground image and its mask are warped by the Spatial Transformer generator $\G$, and the resulting composite is assessed by the discriminator $\D$.
In this setup, $\D$ tries to distinguish warped composites from real images, while $\G$ tries to fool $\D$ by generating as realistic looking as possible composites.
To the best of our knowledge, we are the first to address the problem of realistic image generation through geometric transformations in a GAN framework.
We demonstrate this method on the application of compositing furniture into indoor scenes, which gives a preview of, for example, how purchased items would look in a house.
To evaluate in this domain, we created a synthetic dataset of indoor scene images as the background with masked objects as the foreground.
We also demonstrate ST-GANs in a fully unpaired setting for the task of compositing glasses on portrait images.
A large-scale user study shows that our approach improves the realism of image composites. 

Our main contributions are as follows: 
\begin{itemize}
\setlength\itemsep{0px}
\item We integrate the STN and GAN frameworks and introduce ST-GAN, a novel GAN framework for finding realistic-looking geometric warps. 
\item We design a multi-stage architecture and training strategy that improves warping convergence of ST-GANs.
\item We demonstrate compelling results in image compositing tasks in both paired and unpaired settings as well as its applicability to high-resolution images.
\end{itemize}


\section{Related Work}

\textbf{Image compositing} refers to the process of overlaying a masked foreground image on top of a background image. 
One of the main challenges of image compositing is that the foreground object usually comes from a different scene than the background, and therefore it is not likely to match the background scene in a number of ways that negatively effects the realism of the composite.
These can be both appearance differences (due to lighting, white balance, and shading differences) and geometric differences (due to changes in camera viewpoint and object positioning).

Existing photo-editing software features various image appearance adjustment operations for that allows users to create realistic composites. 
Prior work has attempted to automate appearance corrections (\eg contrast, saturation) through Poisson blending~\cite{perez2003poisson} or more recent deep learning approaches~\cite{zhu2015learning,tsai2017deep}.
In this work, we focus on the second challenge: correcting for \emph{geometric} inconsistencies between source and target images.

\textbf{Spatial Transformer Networks (STNs)}~\cite{jaderberg2015spatial}
are one way to incorporate learnable image warping within a deep learning framework.
A Spatial Transformer module consists of a subnetwork predicting a set of warp parameters followed by a (differentiable) warp function.

STNs have been shown effective in resolving geometric variations for discriminative tasks as well as a wide range of extended applications such as robust filter learning~\cite{dai2017deformable,jia2016dynamic}, image/view synthesis~\cite{zhou2016view,ganin2016deepwarp,park2017transformation,yeh2016semantic}, and 3D representation learning~\cite{kanazawa2016warpnet,yan2016perspective,zhou2017unsupervised}.
More recently, Inverse Compositional STNs (IC-STNs)~\cite{lin2017inverse} advocated an iterative alignment framework.
In this work, we borrow the concept of iterative warping but do not enforce recurrence in the geometric prediction network; instead, we add different generators at each warping step with a sequential training scheme. 

\textbf{Generative Adversarial Networks (GANs)}~\cite{goodfellow2014generative}
are a class of generative models that are learned by playing a minimax optimization game between a generator network $\G$ and a discriminator network $\D$.
Through this adversarial process, GANs are shown to be capable of learning a generative distribution that matches the empirical distribution of a given data collection.
One advantage of GANs is that the loss function is essentially learned by the discriminator network, which allows for training in cases where ground truth data with strong supervision is not available.

GANs are utilized for data generation in various domains, including images~\cite{radford2015unsupervised}, videos~\cite{vondrick2016generating}, and 3D voxelized data~\cite{wu2016learning}.
For images in particular, it has been shown to generate compelling results in a vast variety of conditional image generation problems such as super-resolution~\cite{lai2017deep}, inpainting~\cite{pathak2016context}, image-to-image translation~\cite{isola2016image,zhu2017unpaired,liu2017unsupervised}, and image editing/manipulation~\cite{zhu2016generative}.

Recently, STNs were also sought to be adversarially trained for object detection~\cite{wang2017fast}, where adversarial examples with feature deformations are generated to robustify object detectors.
LR-GAN~\cite{yang2017lr} approached direct image generation problems with additional STNs onto the (directly) generated images to factorize shape variations.
We explore the context of STNs with GANs in the space of \emph{conditional} image generation from given inputs, which is a more direct integration of the two frameworks. 


\section{Approach}

\begin{figure*}[t!]
\centering
\includegraphics[width=.95\linewidth]{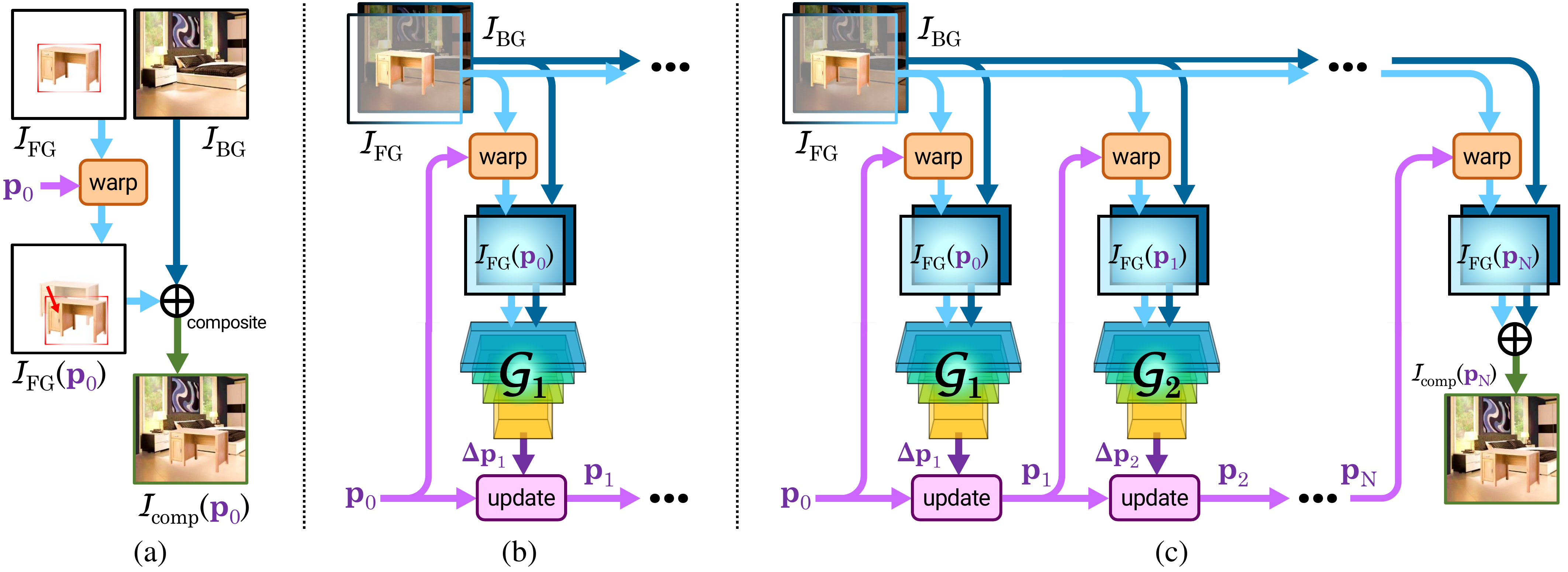}
\caption{Background.
(a) Given an initial composite transformation $\p_0$, the foreground image and mask is composited onto the background image using~\eqref{eq:composite}.
(b) Using \textbf{Spatial Transformer Networks} (STNs), a geometric prediction network $\G_1$ predicts an update $\deltap_1$ conditioned on the foreground and background images, resulting in the new parameters $\p_1$.
The update is performed with warp composition~\eqref{eq:ith}.
(c) Our final form is an \textit{iterative} STN to predict a series of accumulative warp updates on the foreground such that the resulting composite image falls closer to the natural image manifold.
}
\label{fig:composite}
\end{figure*}

Our goal is \textit{realistic geometric correction} for image compositing given a background image $\Ibg$ and foreground object $\Ifg$ with a corresponding mask $\Mfg$.
We aim to correct the camera perspective, position and orientation of the foreground object such that the resulting composite looks natural.
The compositing process can be expressed as:
\begin{align} \label{eq:composite}
\Icomp &= \Ifg \odot \Mfg + \Ibg \odot (1-\Mfg) \nonumber \\
&= \Ifg \oplus \Ibg \;.
\end{align}
For simplicity, we further introduce the notation $\oplus$ to represent compositing (with $\Mfg$ implied within $\Ifg$).
Given the composite parameters $\p_0$ (defining an initial warp state) of $\Ifg$, we can rewrite~\eqref{eq:composite} as
\begin{equation} \label{eq:composite2}
\Icomp(\p_0) = \Ifg(\p_0) \oplus \Ibg \;,
\end{equation}
where images are written as functions of the warp parameters.
This operator is shown in Fig.~\ref{fig:composite}(a).

In this work, we restrict our geometric warp function to homography transformations, which can represent approximate 3D geometric rectifications for objects that are mostly planar or with small perturbations.
As a result, we are making an assumption that the perspective of the foreground object is \emph{close} to the correct perspective; this is often the case when people are choosing similar, but not identical, images from which to composite the foreground object.

The core module of our network design is an STN (Fig.~\ref{fig:composite}(b)), where the geometric prediction network $\G$ predicts a correcting update $\deltap_1$.
We condition $\G$ on both the background and foreground images, since knowing how an object should be transformed to fit a background scene requires knowledge of the complex interaction between the two.
This includes geometry of the object and the background scene, the relative camera position, and semantic understanding of realistic object layouts (\eg having a window in the middle of the room would not make sense).

\subsection{Iterative Geometric Corrections} \label{sec:iterative}

Predicting large displacement warp parameters from image pixels is extremely challenging, so most prior work on image alignment predict local geometric transformations in an iterative fashion~\cite{horn1981determining,lucas1981iterative,baker2004lucas,xiong2013supervised,lin2016conditional}.
Similarly, we propose to use iterative STNs to predict a series of warp updates, shown in Fig.~\ref{fig:composite}(c).
At the $i$th iteration, given the input image $\I$ and the previous warp state $\p_{i-1}$, the correcting warp update $\deltap_i$ and the new warp state $\p_i$ can be written as
\begin{align} \label{eq:ith}
\deltap_i &= \G_i \big( \Ifg(\p_{i-1}),\Ibg \big) \nonumber \\
\p_i &= \p_{i-1} \circ \deltap_i \;\;,
\end{align}
where $\G_i(\cdot)$ is the geometric prediction network and $\circ$ denotes composition of warp parameters.
This family of iterative STNs preserves the original images from loss of information due to multiple warping operations~\cite{lin2017inverse}.

\subsection{Sequential Adversarial Training} \label{sec:adversarial}

\begin{figure}[t!] 
\centering
\includegraphics[width=\linewidth]{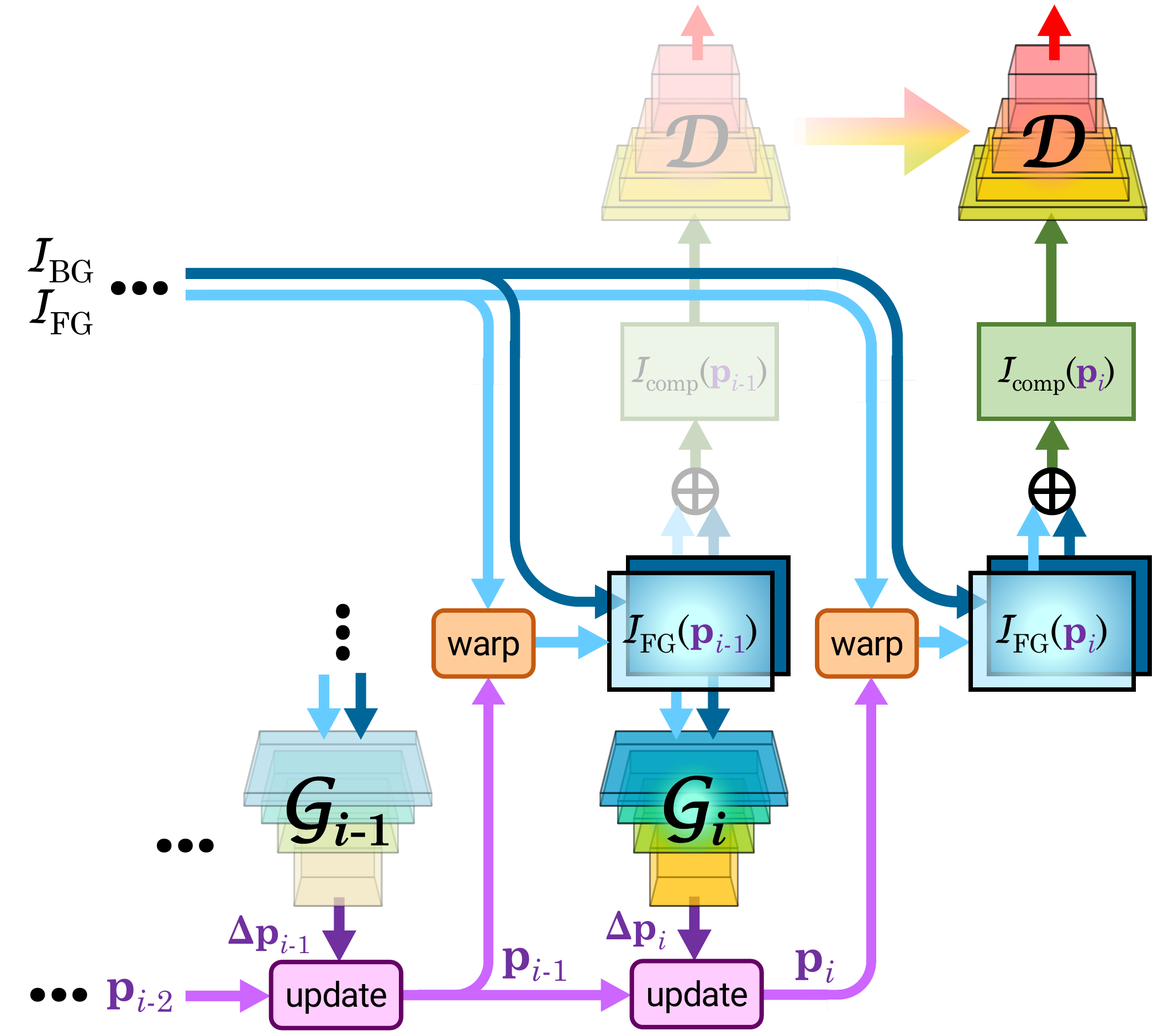}
\caption{\textbf{Sequential adversarial training} of ST-GAN.  When learning a new warp state $\p_i$, only the new generator $\G_i$ is updated while the previous ones are kept fixed.
A single discriminator (learned from all stages) is continuously improved during the sequential learning process.}
\label{fig:adversarial}
\end{figure}

In order for STNs to learn geometric warps that map images closer to the natural image manifold, we integrate them into a GAN framework, which we refer to as ST-GANs.
The motivation for this is two-fold.
First, learning a realistic geometric correction is a multi-modal problem (\eg a bed can reasonably exist in multiple places in a room); second, supervision for these warp parameters are typically not available.
The main difference of ST-GANs from conventional GANs is that (1) $\G$ generates a set of \textit{low-dimensional} warp parameter updates instead of images (the whole set of pixel values); and (2) $\D$ gets as input the warped foreground image composited with the background. 

To learn gradual geometric improvements toward the natural image manifold, we adopt a sequential adversarial training strategy for iterative STNs (Fig.~\ref{fig:adversarial}), where the geometric predictor $\G$ corresponds to the stack of generators $\G_i$.
We start by training a single $\G_1$, and each subsequent new generator $\G_i$ is added and trained by fixing the weights of all previous generators $\{\G_j\}_{j=1 \cdots i-1}$.
As a result, we train only $\G_i$ and $\D$ by feeding the resulting composite image at warp state $\Icomp(\p_i)$ into the discriminator $\D$ and matching it against the real data distribution.
This learning philosophy shares commonalities with the Supervised Descent Method~\cite{xiong2013supervised}, where a series of linear regressors are solved greedily, and we found it makes the overall training faster and more robust.
Finally, we fine-tune the entire network end-to-end to achieve our final result. 
Note that we use the same discriminator $\D$ for all stages of the generator $\G_i$, as the fundamental measure of ``geometric fakeness'' does not change over iterations.

\subsection{Adversarial Objective} \label{sec:objective}

We optimize the Wasserstein GAN (WGAN)~\cite{arjovsky2017wasserstein} objective for our adversarial game.
We note that ST-GAN is amenable to any other GAN variants~\cite{mao2016least,zhao2016energy,berthelot2017began}, and that the choice of GAN architecture is orthogonal to this work.

The WGAN minimax objective at the $i$th stage is
\begin{align} \label{eq:minimax}
\min_{\G_i} \max_{\D \in \mathbb{D}} \; \mathop{\E}_{\substack{x \sim \Pset_\text{fake} \\ \p_i \sim \Pset_{\p_i|\p_{i-1}}}}  \big[ \D \big(x(\p_i) \big) \big]
- \mathop{\E}_{y \sim \Pset_\text{real} } \big[\D(y)\big] \;\;,
\end{align}
where $y = \Ireal$ and $x = \Icomp$ are drawn from the real data and fake composite distributions, and $\mathbb{D}$ is the set of 1-Lipschitz functions enforced by adding a gradient penalty term $\loss_\text{grad}$~\cite{gulrajani2017improved}.
Here, $\p_i$ (where $\G_i$ is implied, defined in~\eqref{eq:ith}) is drawn from the posterior distribution conditioned on $\p_{i-1}$ (recursively implied).
When $i=1$, the initial warp $\p_0$ is drawn from $\Pset_\text{pert}$, a predefined distribution for geometric data augmentation.

We also constrain the warp update $\deltap_i$ to lie within a trust region by introducing an additional penalty $\loss_\text{update} = \| \deltap_i \|_2^2\;$.
This is essential since ST-GAN may learn trivial solutions to remove the foreground (\eg by translating it outside the image or shrinking it into nothing), leaving behind only the background image and in turn making the composite image realistic already.

When training ST-GAN sequentially, we update $\D$ and $\G_i$ alternating the respective loss functions:
\begin{align} \label{eq:lossDG}
\loss_\D &= \E_{x,\p_i} \big[ \D \big(x(\p_i) \big) \big]
- \E_{y} \big[\D(y)\big] 
+ \lambda_\text{grad} \cdot \loss_\text{grad} \\
\loss_{\G_i} &= -\E_{x,\p_i} \big[ \D \big(x(\p_i) \big) \big]
+ \lambda_\text{update} \cdot \loss_\text{update} \;,
\end{align}
where $\lambda_\text{grad}$ and $\lambda_\text{update}$ are the penalty weights for the $\D$ gradient and the warp update $\deltap_i$ respectively, and $\G_i$ and $\deltap_i$ are again implied through~\eqref{eq:ith}. 
When fine-tuning ST-GAN with $N$ learned updates end-to-end, the generator objective is the sum of that from each $\G_i$, \ie $\loss_\G = \sum_{i=1}^N \loss_{\G_i}$ .


\section{Experiments}

We begin by describing the basic experimental settings.

\paragraph{Warp parameterizations.}
We parameterize a homography with the $\mathfrak{sl}(3)$ Lie algebra~\cite{mei2006homography}, \ie the warp parameters $\p \in \mathfrak{sl}(3)$ and homography matrices $\H \in \mathbb{SL}(3)$ are related through the exponential map.
Under this parameterization, warp composition can be expressed as the addition of parameters, \ie $\p_a \circ \p_b \equiv \p_a + \p_b \;\;\; \forall \p_a,\p_b \in \mathfrak{sl}(3)$.

\paragraph{Model architecture.}
We denote the following: {\bf C}($k$) is a 2D convolutional layer with $k$ filters of size $4\times 4$ and stride 2 (halving the feature map resolution) and {\bf L}($k$) is a fully-connected layer with $k$ output nodes.
The input of the generators $\G_i$ has 7 channels: RGBA for foreground and RGB for background, and the input to the discriminator $\D$ is the composite image with 3 channels (RGB).
All images are rescaled to $120\times 160$, but we note that the parameterized warp can be applied to full-resolution images at test time.

The architecture of $\G$ is {\bf C}(32)-{\bf C}(64)-{\bf C}(128)-{\bf C}(256)-{\bf C}(512)-{\bf L}(256)-{\bf L}(8), where the output is the 8-dimensional (in the case of a homography) warp parameter update $\deltap$.
For each convolutional layer in $\G$, we concatenate a downsampled version of the original image (using average pooling) with the input feature map.
For $\D$, we use a PatchGAN architecture~\cite{isola2016image}, with layout {\bf C}(32)-{\bf C}(64)-{\bf C}(128)-{\bf C}(256)-{\bf C}(512)-{\bf C}(1).
Nonlinearity activations are inserted between all layers, where they are ReLU for $\G$ and LeakyReLU with slope 0.2 for $\D$.
We omit all normalization layers as we found them to deteriorate training performance.

\subsection{3D Cubes}

\begin{figure}[t!] \center
\includegraphics[width=0.99\linewidth]{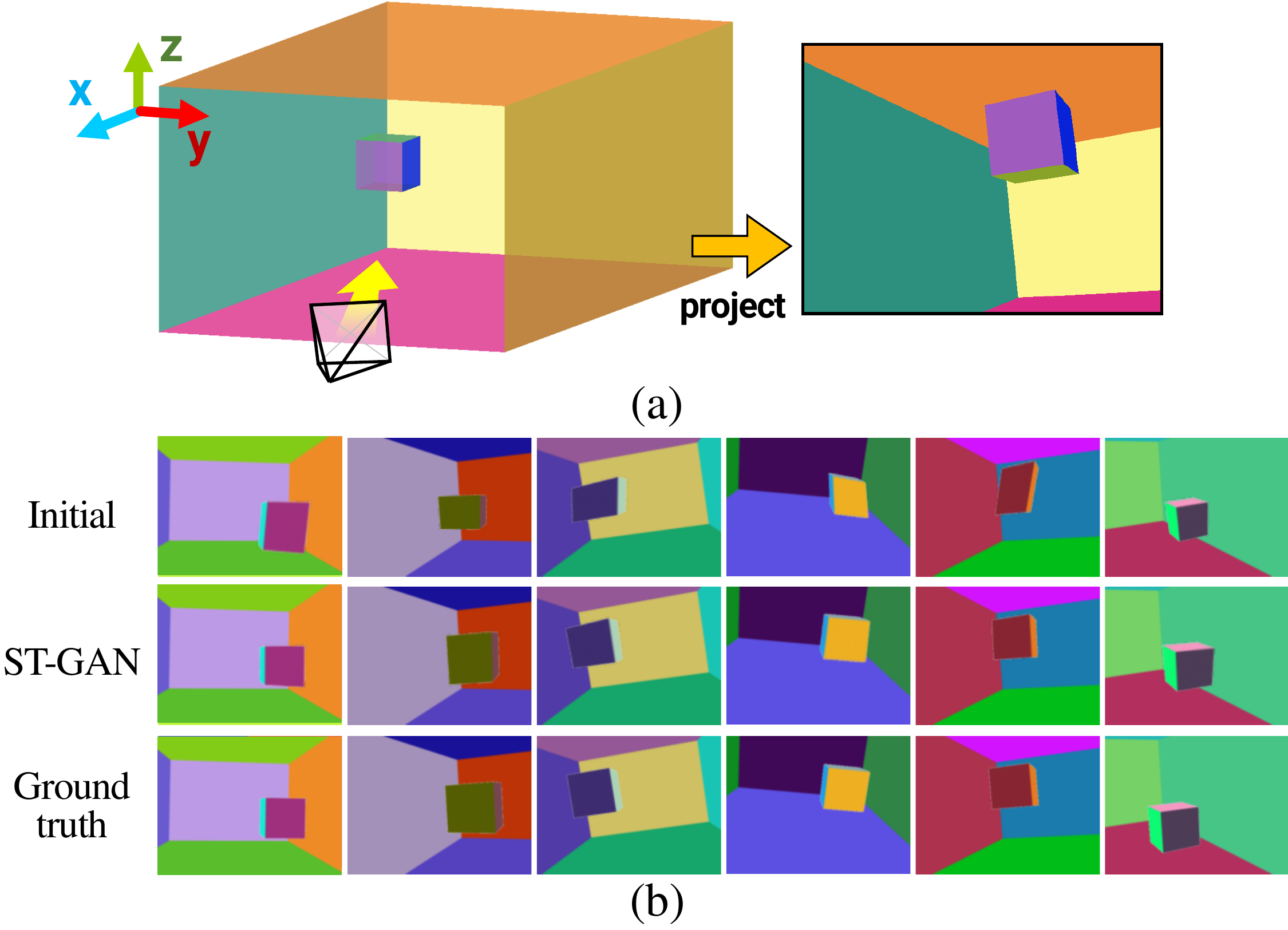}
\caption{(a) We create a synthetic dataset of 3D cube renderings and validate the efficacy of ST-GAN by attempting to correct randomly generated geometric perturbations.
(b) ST-GAN is able to correct the cubes to a right perspective, albeit a possible translational offset from the ground truth.
}
\label{fig:cubes}
\end{figure}

To begin with, we validate whether ST-GANs can make geometric corrections in a simple, artificial setting.
We create a synthetic dataset consisting of a 3D rectangular room, an axis-aligned cube inside the room, and a perspective camera (Fig.~\ref{fig:cubes}(a)).
We apply random 3-DoF translations to the cube and 6-DoF perturbations to the camera, and render the cube/room pair separately as the foreground/background (of resolution $120\times 160$).
We color all sides of the cube and the room randomly.

We perturb the rendered foreground cubes with random homography transformations as the initial warp $\p_0$ and train ST-GAN by pairing the original cube as the ground-truth counterpart for $\D$.
As shown in Fig.~\ref{fig:cubes}(b), ST-GAN is able to correct the perturbed cubes scale and perspective distortion \wrt the underlying scene geometry.
In addition, ST-GAN is sometimes able to discover other realistic solutions (\eg not necessarily aligning back to the ground-truth location), indicating ST-GAN's ability to learn the multi-modal distribution of correct cube placements in this dataset.

\subsection{Indoor Objects}

Next, we show how ST-GANs can be applied to practical image compositing domains.
We choose the application of compositing furniture in indoor scenes and demonstrate its efficacy on both simulated and real-world images.
To collect training data, we create a synthetic dataset consisting of rendered background scenes and foreground objects with masks.
We evaluate on the synthetic test set as well as high-resolution real world photographs to validate whether ST-GAN also generalizes to real images.

\begin{table}[t!]
\rowcolors{3}{white}{myblue!10}
\resizebox{.97\linewidth}{!}{
\setlength{\tabcolsep}{10px}
\begin{tabular}{r r r r r}
\toprule
\multicolumn{1}{r}{\multirow{2}{*}{\bf Category}} & \multicolumn{2}{c}{\bf Training set} & \multicolumn{2}{c}{\bf Test set} \\
& \# 3D inst. & \# pert. & \# 3D inst. & \# pert.  \\
\midrule
Bed          & 3924	& 11829	& 414	& 1281	\\
Bookshelf    & 508	& 1280	& 58	& 137	\\
Cabinet      & 9335	& 31174	& 1067	& 3518	\\
Chair        & 196	& 609	& 22	& 60	\\
Desk         & 64	& 1674	& 73	& 214	\\
Dresser      & 285	& 808	& 31	& 84	\\
Refrigerator & 3802	& 15407	& 415	& 1692	\\
Sofa         & 3604	& 11165	& 397	& 1144	\\
\midrule
\multicolumn{1}{r}{\bf Total}  & 22303 & 73946 & 2477 & 8130 \\
\bottomrule
\end{tabular}
}
\caption{Dataset statistics for the indoor object experiment, reporting the number of object instances chosen for perturbation, and the final number of rendered perturbed samples.}
\label{table:suncg_data}
\end{table}

\paragraph{Data preparation.}
We render synthetic indoor scene images from the SUNCG dataset~\cite{song2016ssc}, consisting of 45,622 indoor scenes with over 5M 3D object instances from 37 categories~\cite{silberman2012indoor}.
We use the selected 41,499 scene models and the 568,749 camera viewpoints from Zhang~\etal~\cite{zhang2016physically} and utilize Mitsuba~\cite{Mitsuba} to render photo-realistic images with global illumination.
We keep a list of candidate 3D objects consisting of all instances visible from the camera viewpoints and belonging to the categories listed in Table~\ref{table:suncg_data}.

\begin{figure}[t!] \center
\includegraphics[width=.95\linewidth]{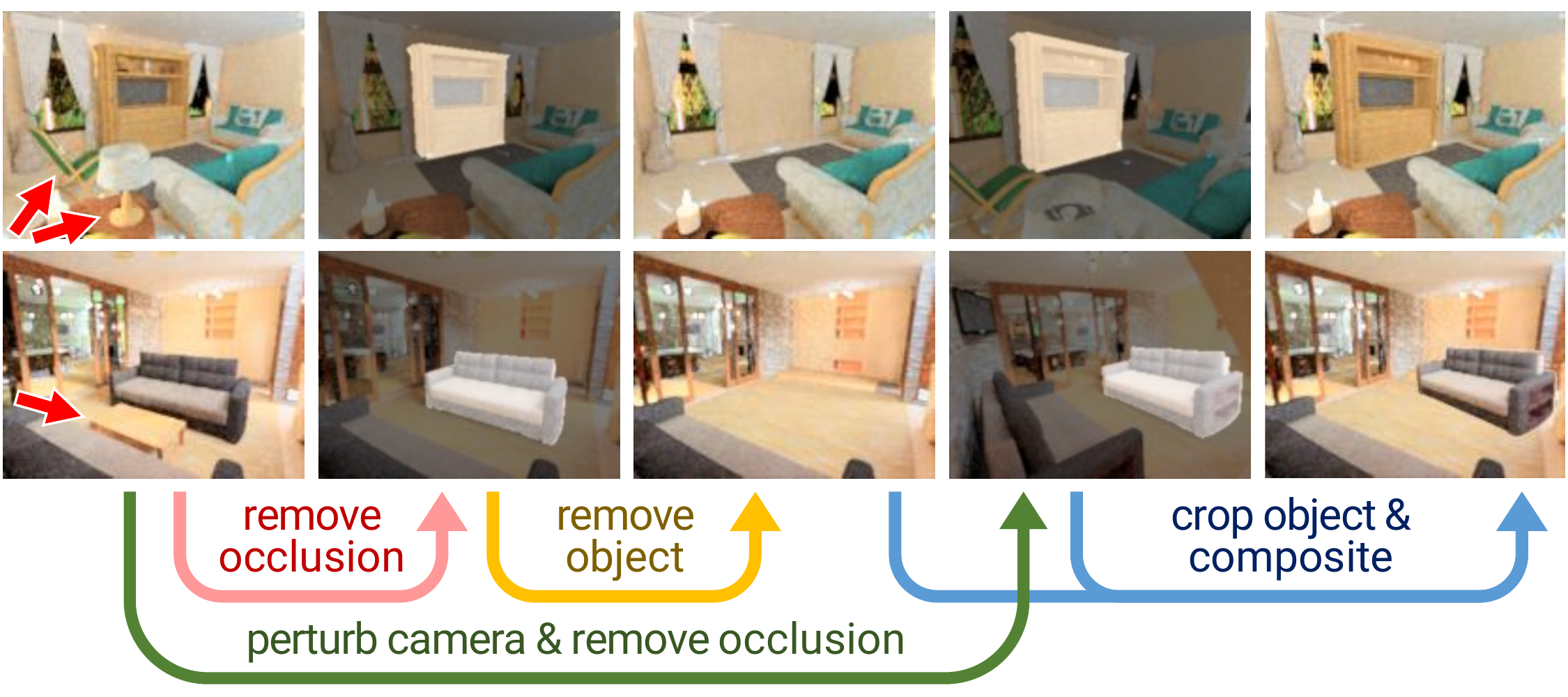}
\caption{\textbf{Rendering pipeline.}
Given an indoor scene and a candidate object, we remove occluding objects to create an occlusion-free scenario, which we do the same at another perturbed camera pose.
We further remove the object to create a training sample pair with mismatched perspectives.
}
\label{fig:suncg_render}
\end{figure}

\begin{table*}[t]
	\rowcolors{3}{white}{myblue!10}
	\resizebox{.97\linewidth}{!}{
		\begin{tabular}{r c c c c c c c c|c}		
			\toprule
			\multicolumn{1}{r}{\multirow{2}{*}{\bf Category}} & Initial     & \multirow{2}{*}{SDM~\cite{xiong2013supervised}} & Homogra- & ST-GAN & ST-GAN & ST-GAN & ST-GAN & \bf ST-GAN & Ground \\
			
			& warp      &          & phyNet~\cite{detone2016deep}  & (non-seq.)  & (warp 1)    & (warp 2)      & (warp 4)      & \bf (end-to-end)  & truth \\
			\midrule
			Bed         & 35.5 \%  & 30.5 \% & 30.2 \%     & 32.8 \%    & 32.8 \%    & \bf 46.8 \%  & 32.8 \%      & 32.2 \%      & 75.0 \% \\
			Bookshelf   & 21.1 \%  & 33.9 \% & 35.1 \%     & 16.7 \%    & 26.4 \%    & 26.2 \%      & 39.5 \%      & \bf 42.6 \%  & 68.9 \% \\
			Cabinet     & 20.9 \%  & 19.8 \% & 35.0 \%     & 36.6 \%    & 14.3 \%    & 31.2 \%      & 44.4 \%      & \bf 50.0 \%  & 74.3 \% \\
			Chair       & 32.8 \%  & 36.8 \% & 47.6 \%     & 50.9 \%    & \bf 62.3\% & 42.7 \%      & 50.0 \%      & 58.6 \%      & 68.7 \% \\
			Desk        & 18.9 \%  & 13.1 \% & 36.1 \%     & 35.4 \%    & 29.2 \%    & 29.0 \%      & 39.4 \%      & \bf 40.7 \%  & 65.1 \% \\
			Dresser     & 14.9 \%  & 18.6 \% & 20.7 \%     & 16.7 \%    & 24.6 \%    & 27.4 \%      & 29.7 \%      & \bf 48.4 \%  & 66.1 \% \\
			Refrigerator & 37.1 \% & 21.4 \% & 50.0 \%     & 37.7 \%    & 28.6 \%    & 47.1 \%      & 39.7 \%      & \bf 51.7 \%  & 81.6 \% \\
			Sofa        & 15.9 \%  & 31.0 \% & 42.4 \%     & 28.9 \%    & 37.0 \%    & 54.9 \%      & \bf 56.1 \%  & 51.8 \%      & 78.2 \% \\
			\midrule
			\multicolumn{1}{c}
			{\bf Average}& 24.6 \% & 25.6 \% & 37.1 \%     & 31.9 \%    & 31.9 \%    & 38.2 \%      & 41.5 \%      & \bf47.0 \%   & 72.6 \% \\
			\bottomrule
			
		\end{tabular}
	}
	\caption{\textbf{AMT User studies} for the indoor objects experiment.
		Percentages represent the how often the images in each category were classified as ``real'' by Turkers.
		We can see that our final model, ST-GAN (end-to-end), substantially improves over geometric realism when averaged across all classes.
		Our realism performance improves with the number of warps trained as well as after the end-to-end fine-tuning.
		The ground truth numbers serve as a theoretical upper bound for all methods.}
	\label{table:AMT}
\end{table*}

The rendering pipeline is shown in Fig.~\ref{fig:suncg_render}.
During the process, we randomly sample an object from the candidate list, with an associated camera viewpoint.
To emulate an occlusion-free compositing scenario, occlusions are automatically removed by detecting overlapping object masks.
We render one image with the candidate object present (as the ``real'' sample) and one with it removed (as the background image).
In addition, we perturb the 6-DoF camera pose and render the object with its mask (as the foreground image) for compositing.
We thus obtain a rendered object as viewed from a \emph{different camera perspective}; this simulates the image compositing task where the foreground and background perspectives mismatch.
We note that a homography correction can only approximate these 3D perturbations, so there is \emph{no planar ground-truth warp} to use for supervision.
We report the statistics of our rendered dataset in Table~\ref{table:suncg_data}.
All images are rendered at $120\times 160$ resolution.

\paragraph{Settings.}
Similar to the prior work by Lin \& Lucey~\cite{lin2017inverse}, we train ST-GAN for $N=4$ sequential warps
During adversarial training, we rescale the foreground object randomly from $\text{Unif}(0.9,1.1)$ and augment the initial warp $\p_0$ with a translation sampled from $\mathcal{N}(0,0.05)$ scaled by the image dimensions.
We set $\lambda_\text{update}=0.3$ for all methods.

\paragraph{Baselines.}
One major advantage of ST-GAN is that it can learn from ``realism'' comparisons \emph{without ground-truth warp parameters} for supervision.
However, prior approaches require supervision directly on the warp parameters.
Therefore, we compare against self-supervised approaches trained with random homography perturbations on foreground objects as input, yielding warp parameters as self-supervision.
We reemphasize that such direct supervision is insufficient in this application as we aim to find the closest point on a manifold of realistic looking composites rather than fitting a specific paired model.
Our baselines are (1) HomographyNet~\cite{detone2016deep}, a CNN-based approach that learns direct regression on the warp parameters, and (2) Supervised Descent Method (SDM)~\cite{xiong2013supervised}, which greedily learns the parameters through cascaded linear regression.
We train the SDM baseline for 4 sequential warps as well.

\paragraph{Quantitative evaluation.}
As with most image generation tasks where the goal is realism, there is no natural quantitative evaluation possible.
Therefore, we carry out a perceptual study on Amazon Mechanical Turk (AMT) to assess geometric realism of the warped composites.
We randomly chose 50 test images from each category and gather data from 225 participants.
Each participant was shown a composite image from a randomly selected algorithm (Table~\ref{table:AMT}), and was asked whether they saw any objects whose shape does not look natural in the presented image.

We report the AMT assessment results in Table~\ref{table:AMT}.
On average, ST-GAN shows a large improvement of geometric realism, and quality improves over the sequential warps.
When considering that the warp is restricted to homography transformations, these results are promising, as we are not correcting for more complicated view synthesis effects for out-of-plane rotations such as occlusions.
Additionally, ST-GAN, which does not require ground truth warp parameters during training, greatly outperforms other baselines, while SDM yields no improvement and HomographyNet increases realism, but to a lesser degree.

\paragraph{Ablation studies.}
We found that learning iterative warps is advantageous:
compared with a non-iterative version with the same training iterations (non-seq. in Table~\ref{table:AMT}), ST-GAN (with multiple generators) approaches geometric realism more effectively with iterative warp updates.
In addition, we trained an iterative HomographyNet~\cite{detone2016deep} using the same sequential training strategy as ST-GAN but found little visual improvement over the non-iterative version; we thus focus our comparison against the original~\cite{detone2016deep}.

\begin{figure*}[t!] \center
\includegraphics[width=\linewidth]{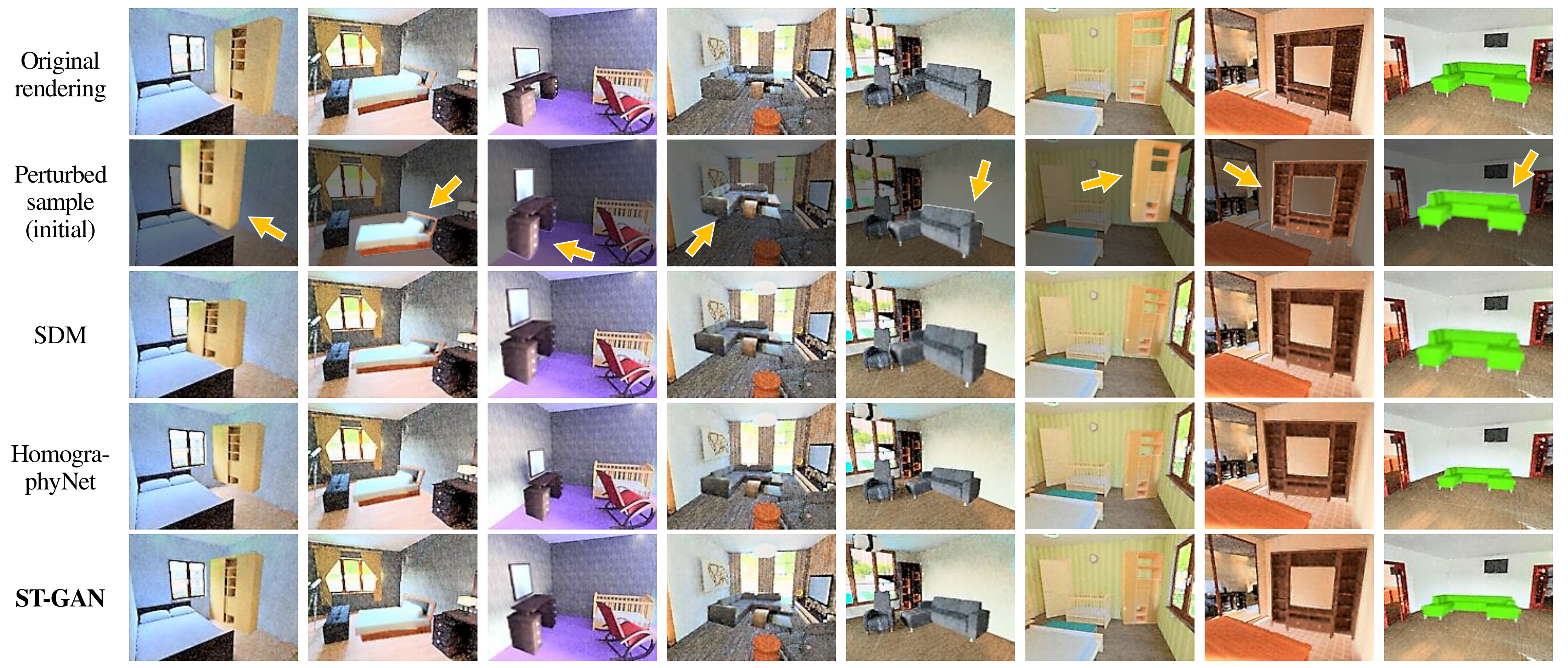}
\caption{\textbf{Qualitative evaluation} on the indoor rendering test set.
Compared to the baselines trained with direct homography supervision, ST-GAN creates more realistic composites.
We find that ST-GAN is able to learn common object-room relationships in the dataset, such as beds being against walls.
Note that ST-GANs corrects the perspectives but not necessarily scale, as objects often exist at multiple scales in the real data. 
We observe that ST-GAN occasionally performs worse for unusual objects (\eg with peculiar colors, last column).
}
\label{fig:suncg_results}
\end{figure*}

\begin{figure}[t!] \center
\includegraphics[width=1.0\linewidth]{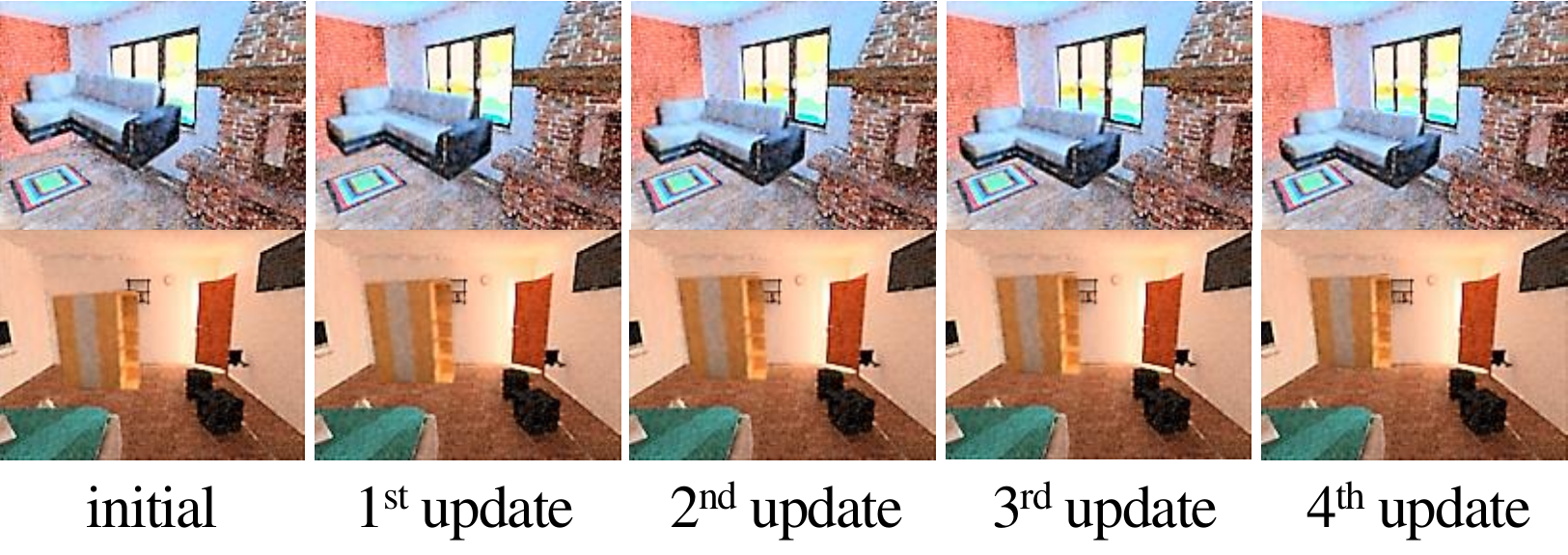}
\caption{Visualization of iterative updates in ST-GAN, where objects make gradual improvements that reaches closer to realism in an incremental fashion.}
\label{fig:gradual}
\end{figure}

\begin{figure}[t!] \center
\includegraphics[width=1.0\linewidth]{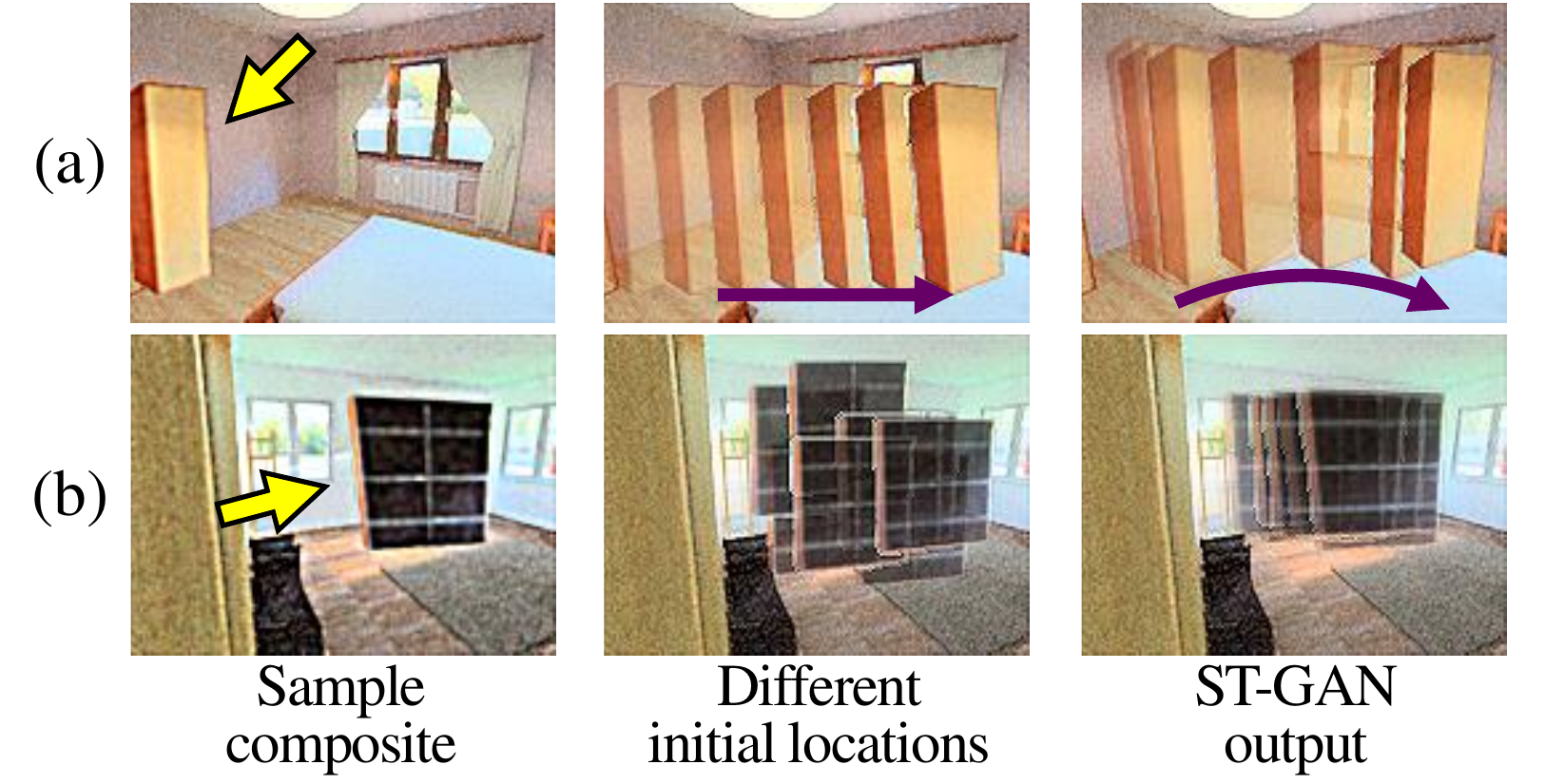}
\caption{\textbf{Dragging and snapping.}
(a) When an object is dragged across the scene, the perspective changes with the composite location to match that of the camera's.
(b) ST-GAN ``snaps'' objects to where it would be frequently composited (\eg a bookshelf is usually laid against the wall).
}
\label{fig:dragsnap}
\end{figure}

\paragraph{Qualitative evaluation.}
We present qualitative results in Fig.~\ref{fig:suncg_results}.
ST-GAN visually outperforms both baselines trained with direct homography parameter supervision, which is also reflected in the AMT assessment results.
Fig.~\ref{fig:gradual} shows how ST-GAN updates the homography warp with each of its generators; we see that it learns gradual updates that makes a realism improvement at each step.
In addition, we illustrates in Fig.~\ref{fig:dragsnap} the effects ST-GAN learns, including gradual changes of the object perspective at different composite locations inside the room, as well as a ``snapping'' effect that predicts a most likely composite location given a neighborhood of initial locations.
These features are automatically learned from the data, and they can be useful when implemented in interactive settings.

\begin{figure}[t!] \center
\includegraphics[width=\linewidth]{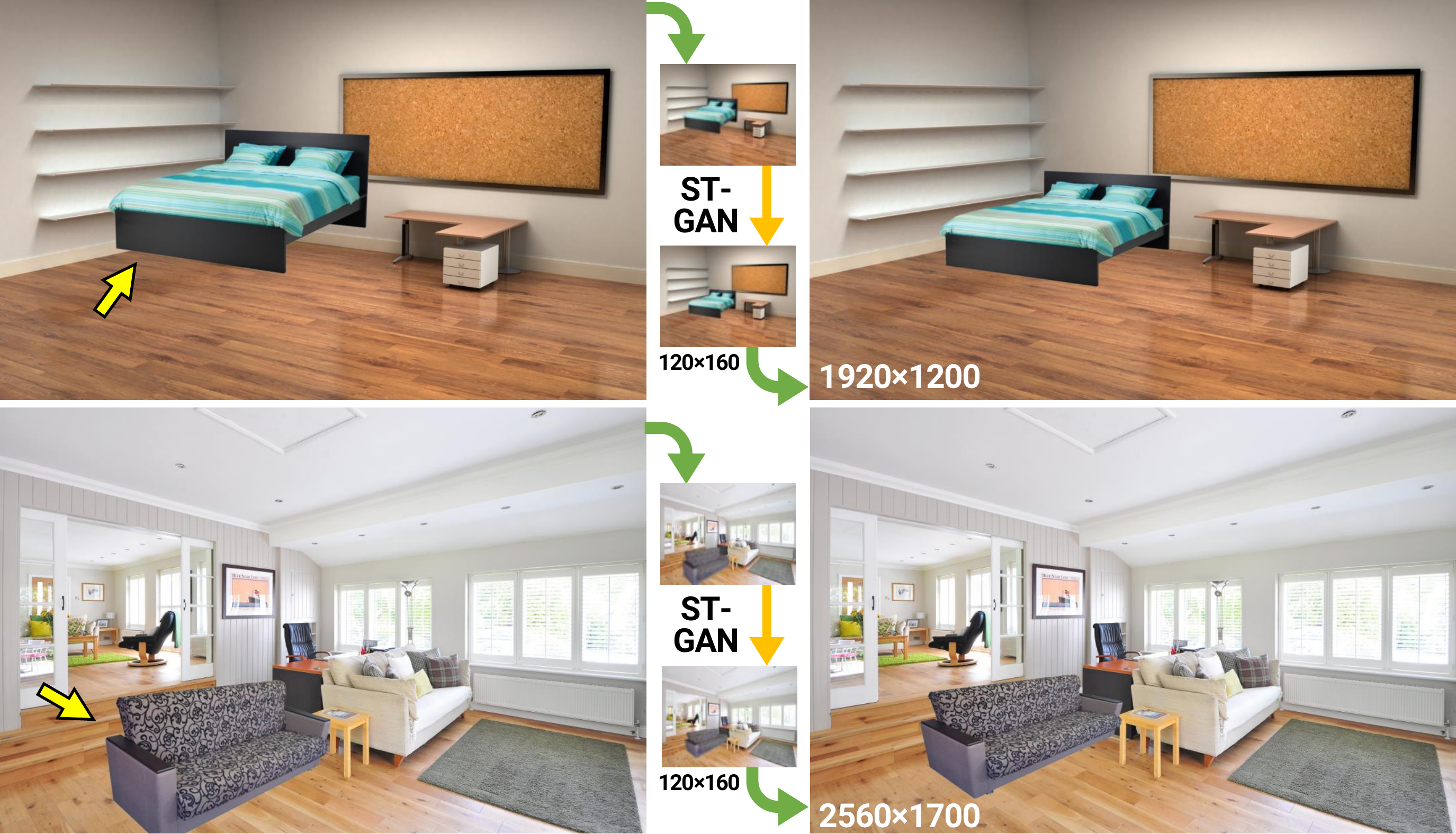}
\caption{\textbf{Real world high-resolution test results.} Here we show our method applied to real images. The inputs are scaled down and fed to the network and then the warp parameters are applied at full resolution.}
\label{fig:suncg_highres}
\end{figure}

Finally, to test whether ST-GAN extends to real images, we provide a qualitative evaluation on photographic, high-resolution test images gathered from the Internet and manually masked (Fig~\ref{fig:suncg_highres}).
This is feasible since the warp parameters predicted from the low-resolution network input are transferable to high-resolution images.
As a consequence, ST-GAN is indirectly applicable to various image resolutions and not strictly limited as with conventional GAN frameworks.
Our results demonstrates the utilization of ST-GAN for high-quality image generation and editing.

\subsection{Glasses}

Finally, we demonstrate results in an entirely unpaired setting where we learn warping corrections for compositing glasses on human faces.
The lack of paired data means that we do not necessarily have pictures of the same people both with and without glasses (ground truth).

\begin{figure}[t!] \center
\includegraphics[width=\linewidth]{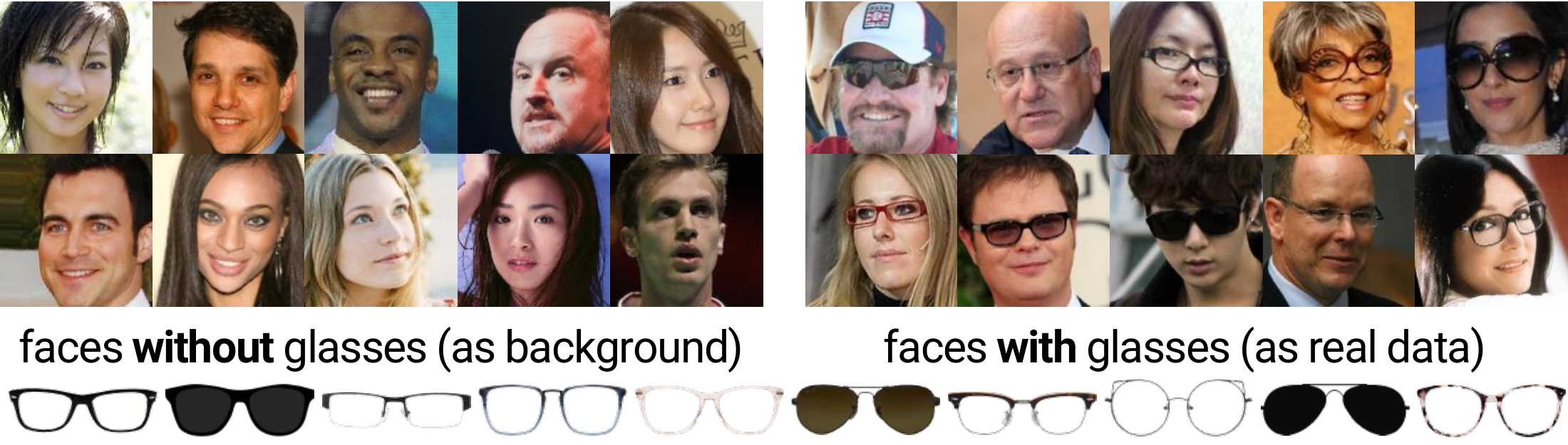}
\caption{The split of CelebA for the background and the real images, as well as the crafted glasses as the foreground.}
\label{fig:glasses}
\end{figure}

\paragraph{Data preparation.}
We use the CelebA dataset~\cite{liu2015faceattributes} and follow the provided training/test split.
We then use the ``eyeglasses'' annotation to separate the training set into two groups.
The first group of people with glasses serve as the real data to be matched against in our adversarial settings, and the group of people without glasses serves as the background.
This results in 152249 training and 18673 test images without glasses, and 10521 training images with glasses.
We hand-crafted 10 pairs of frontal-facing glasses as the foreground source (Fig.~\ref{fig:glasses}).
We note that there are no annotations about where or how the faces are placed, and we do not have any information where the different parts of the glasses are in the foreground images.

In this experiment, we train ST-GAN with $N=5$ sequential warps.
We crop the aligned faces into $144\times 144$ images and resize the glasses to widths of $120$ pixels initialized at the center.
During training, we add geometric data augmentation by randomly perturbing the faces with random similarity transformations and the glasses with random homographies.

\begin{figure}[t!] \center
\includegraphics[width=\linewidth]{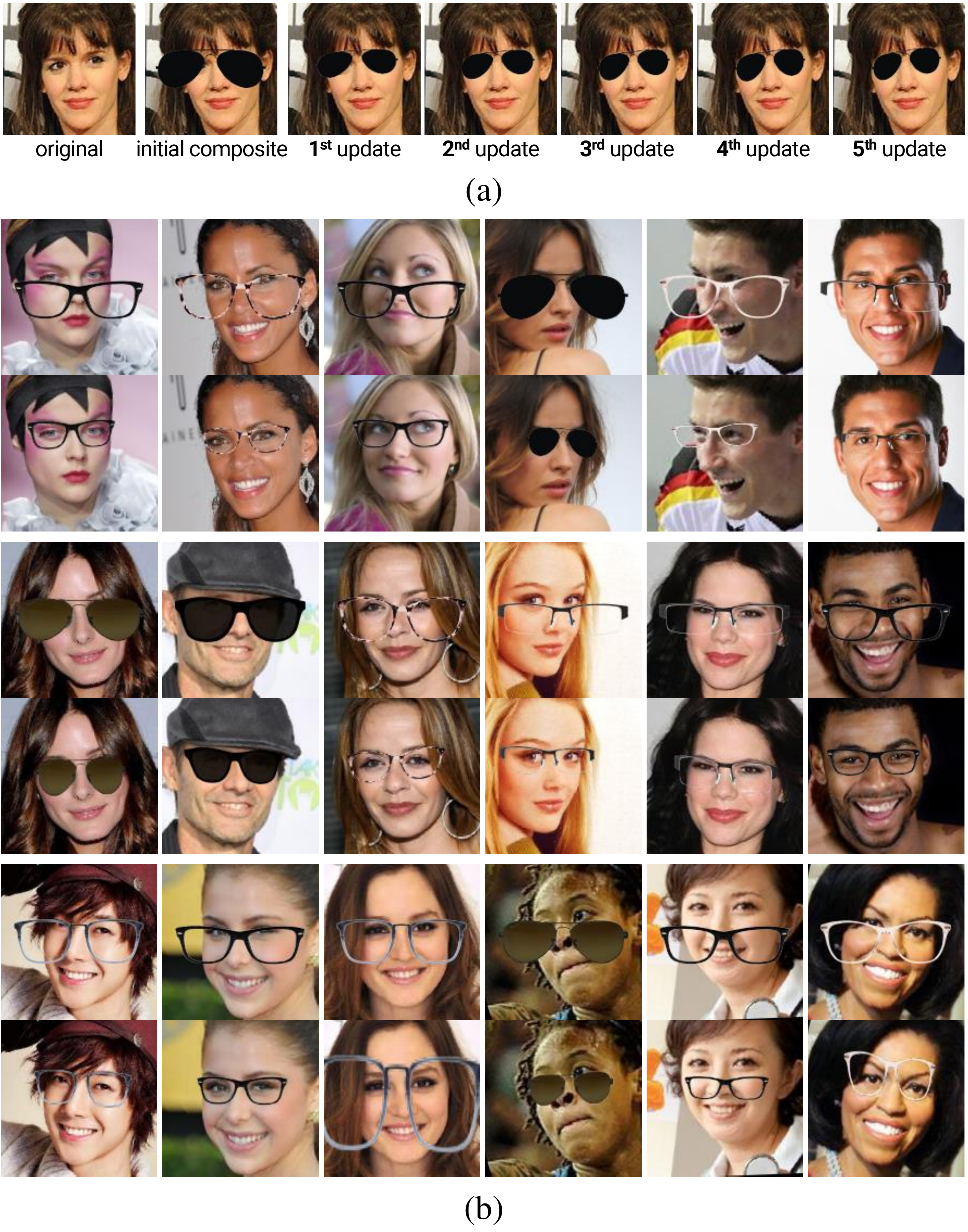}
\caption{\textbf{Glasses compositing results.}
(a) The glasses progressively moves into a more realistic position.
(b) ST-GAN learns to warp various kinds of glasses such that the resulting positions are usually realistic. 
The top rows indicates the initial composite, and the bottom rows indicates the ST-GAN output.
The last 4 examples shows failure cases, where glasses fail to converge onto the faces.
}
\label{fig:glasses_results}
\end{figure}

\paragraph{Results.}
The results are shown in Fig.~\ref{fig:glasses_results}.
As with the previous experiments, ST-GAN learns to warp the foreground glasses in a gradual fashion that improves upon realism at each step.
We find that our method can correctly align glasses onto the people's faces, even with a certain amount of in-plane rotations.
However, ST-GAN does a poorer job on faces with too much out-of-plane rotation.

While such an effect is possible to achieve by taking advantage of facial landmarks, our results are encouraging as no information was given about the structure of either domain, and we only had access to unpaired images of people with and without glasses.
Nonetheless, ST-GAN was able to learn a realism manifold that drove the Spatial Transformer generators.
We believe this demonstrates great potential to extend ST-GANs to other image alignment tasks where acquiring paired data is very challenging.


\section{Conclusion}

We have introduced ST-GANs as a class of methods to model \emph{geometric realism}. 
We have demonstrated the potential of ST-GANs on the task of image compositing, showing improved realism in a large-scale rendered dataset, and results on fully unpaired real-world image data.
It is our hope that this work will open up new revenues to the research community to continue to explore in this direction.

Despite the encouraging results ST-GAN achieves, there are still some limitations.
We find that ST-GAN suffers more when presented imbalanced data, particularly rare examples (\eg white, thick-framed glasses in the glasses experiment).
In addition, we also find convergence of ST-GAN to fail with more extreme translation or in-plane rotation of objects.
We believe a future analysis of the convergence properties of classical image alignment methods with GAN frameworks is worthy of investigation in improving the robustness of ST-GANs.

{\small
\bibliographystyle{ieee}
\bibliography{reference}
}

\section*{Appendix}


\subsection*{A.1. Indoor Object Experiment: Rendering Details}

We describe additional details regarding the rendering of the SUNCG dataset~\cite{song2016ssc} for our experiment.
In addition to Mitsuba~\cite{Mitsuba} for rendering photo-realistic textures, we also utilize the OpenGL toolbox provided by Song~\etal~\cite{song2016ssc}, which supports rendering of instance segmentation.

\paragraph{Candidate object selection.}
For each of the provided camera viewpoints from Zhang~\etal~\cite{zhang2016physically}, we render an instance segmentation of all objects visible in the camera viewpoint.
For each of these objects, we also separately render a binary object mask by removing all other existing objects (including the floor/ceiling/walls).

We use these information to exclude objects that are not ideal for our compositing experiment, including those that are too tiny or only partially visible in the camera view.
Therefore, we include objects into the candidate selection list that match the criteria:
\begin{itemize}
\item The entire object mask is visible within the camera.
\item The object mask occupies at least 10\% of all pixels.
\item At least 50\% of the object mask is visible within the instance segmentation mask.
\item The object belongs to one of the NYUv2~\cite{silberman2012indoor} categories of refrigerators, desks, bookshelves, cabinets, beds, dressers, sofas, or chairs.
\end{itemize}

\paragraph{Occlusion removal.}
For all the objects in the candidate list, we remove the occluding objects (from the associated camera viewpoint) by overlapping the object mask onto the instance segmentation mask.
All overlapped pixels with different instance labels are detected to be associated with an occluding object.
Since there may be ``hidden'' occlusions that are occluded in the first place, we repeat the same process after the initial detected occlusions are removed to reveal the remaining occlusions.
This is repeated until no more occluding objects \wrt the candidate object is present.

In order to create a cleaner space for compositing objects, we also use a ``thicker'' object mask for the above removal procedure.
To achieve this, we dilate the object mask with a $3\times 3$ all-ones kernel for 10 times (\ie ``thicken'' the object mask by 10 pixels).

\paragraph{Camera perturbation.}
For each of the provided camera viewpoints, we generate a camera perturbation by adding a random 3D-translation sampled from $\text{Unif}(-1,1)$ in the forward-backward direction, one sampled from $\text{Unif}(-1,1)$ in the left-right direction (both scaled in meters as defined in the dataset), and a random azimuth rotation sampled from $\text{Unif}(-30,30)$ (degrees).

After generating a camera perturbation, the same occlusion removal process described above is performed to ensure the wholeness of the object from the perturbed perspective.
The candidate object rendered from the perturbed view serves as the foreground source for our experiment.
However, if it becomes only partially or not visible, then the rendering is discarded.

\paragraph{Rendering.}
We use Mitsuba to render $120\times 160$ realistic textures and the OpenGL toolbox to render object masks at $240\times 320$ followed by $\times 2$ downscaling for anti-aliasing.

\subsection*{A.2. Warp Parameterization Details}
We follow Mei~\etal~\cite{mei2006homography} to parameterize homography with the $\mathfrak{sl}(3)$ Lie algebra.
Given a warp parameter vector $\p = [p_1,p_2,p_3,p_4,p_5,p_6,p_7,p_8]^\top \in \mathfrak{sl}(3)$, the transformation matrix $\H \in \mathbb{SL}(3)$ can be written as
\begin{equation}
\H(\p) = \exp \left( 
\begin{bmatrix} p_1 & p_2 & p_3 \\ p_4 & -p_1-p_8 & p_5 \\ p_6 & p_7 & p_8 \end{bmatrix}
\right) \;,
\end{equation}
where $\exp$ is the exponential map (\ie matrix exponential).
$\H$ is the identity transformation when $\p$ is an all-zeros vector.
Warp composition can thus be expressed as the addition of parameters, \ie $\p_a \circ \p_b \equiv \p_a + \p_b \;\; \forall \p_a,\p_b \in \mathfrak{sl}(3)$; furthermore, $\det(\H)=1 \;\; \forall \H \in \mathbb{SL}(3)$.

The exponential map is also Taylor-expandable as
\begin{equation}
\H(\p) = \exp(\X(\p)) = \lim_{K \to \infty} \sum_{k=0}^K \frac{\X^k(\p)}{k!} \;.
\end{equation}
We implement the $\mathfrak{sl}(3)$ parameterization using the Taylor approximation expression with $K=20$.

\subsection*{A.3. Training Details}

For all experiments, we set the batch size for all experiments to be 20.
Unless otherwise specified, we initialize all learnable weights in the networks from $\mathcal{N}(0,0.01)$ and all biases to be 0.
All deep learning approaches are trained with Adam optimization~\cite{kingma2014adam}.
We set $\lambda_\text{grad}=10$ following Gulrajani~\etal~\cite{gulrajani2017improved}.

We describe settings for specific experiments as follows.

\paragraph{3D cubes.}
We create 4000 samples of 3D cube/room pairs with random colors, as described in the paper.
For the initial warp $\p_0$, we generate random homography perturbations $\p_0$ by sampling each element of $\p_0$ from $\mathcal{N}(0,0.1)$, \ie $\p_0 \sim \mathcal{N}(\0,0.1 \mathbf{I})$.
This is applied to a canonical frame with $x$ and $y$ coordinates normalized to $[-1,1]$ and subsequently transformed back to the image frame.
We train ST-GAN with 4 sequential warps, each for 50K iterations (with perturbations generated on the fly) with the learning rates for both $\G$ and $\D$ to be $10^{-4}$.
We set $\lambda_\text{update}=0.1$ in this experiment.

\paragraph{Indoor objects.}
For the self-supervised baselines (HomographyNet~\cite{detone2016deep} and SDM~\cite{xiong2013supervised}), we generate random homography perturbations $\p_0$ using the same noise model as that from the 3D cubes experiment.

We train HomographyNet for 200K iterations (with perturbations generated on the fly) with a learning rate of $10^{-4}$.
For SDM, we vectorize the grayscale images to be the feature as was practiced for image alignment~\cite{lin2016conditional}; in our case, we concatenate those of the background and masked foreground as the final extracted feature.
We generate 750K perturbed examples (more than 10 perturbed examples per training sample) to train each linear regressor.
Also as was practiced~\cite{xiong2013supervised,lin2016conditional}, we add an $\ell_2$ regularization term to the SDM least-squares objective function and search for the penalty factor by evaluating on a separate validation set.

We initialize each of the ST-GAN generators $\G_i$ with the pretrained HomographyNet as we find it to be better-conditioned.
During adversarial training, we train each $\G_i$ for 40K iterations with the learning rate for $\G_i$ to be $10^{-6}$ and that of $\D$ to be $10^{-4}$.
In the final end-to-end fine-tuning stage, we train all $\G_i$ for 40K iterations using the same learning rates  ($10^{-6}$ for all $\G_i$ and $10^{-4}$ for $\D$).
The non-sequential ST-GAN baseline is trained for 160K iterations with the same learning rates.
We set $\lambda_\text{update}=0.3$ in this experiment.

\paragraph{Glasses.}
For data augmentation, we perturb the faces with random similarity transformations from $\mathcal{N}(0,0.1)$ for rotation (radian) and $\mathcal{N}(0,0.05)$ for translation (scaled by the image dimensions, in both $x$ and $y$ directions).
The glasses are perturbed using the same random homography noise model as used in the 3D cubes experiment.

We train ST-GAN with 5 sequential warps, each for 50K iterations with the learning rates for both $\G$ and $\D$ to be $10^{-5}$.
As a preconditioning step, we also pretrain the discriminator $\D$ using only the initial fake samples and real samples for 50K iterations with the same learning rate.
We set $\lambda_\text{update}=1$ in this experiment.

\subsection*{A.4. Additional Indoor Object Results}

\begin{figure*}[t!] \center
\includegraphics[width=\linewidth]{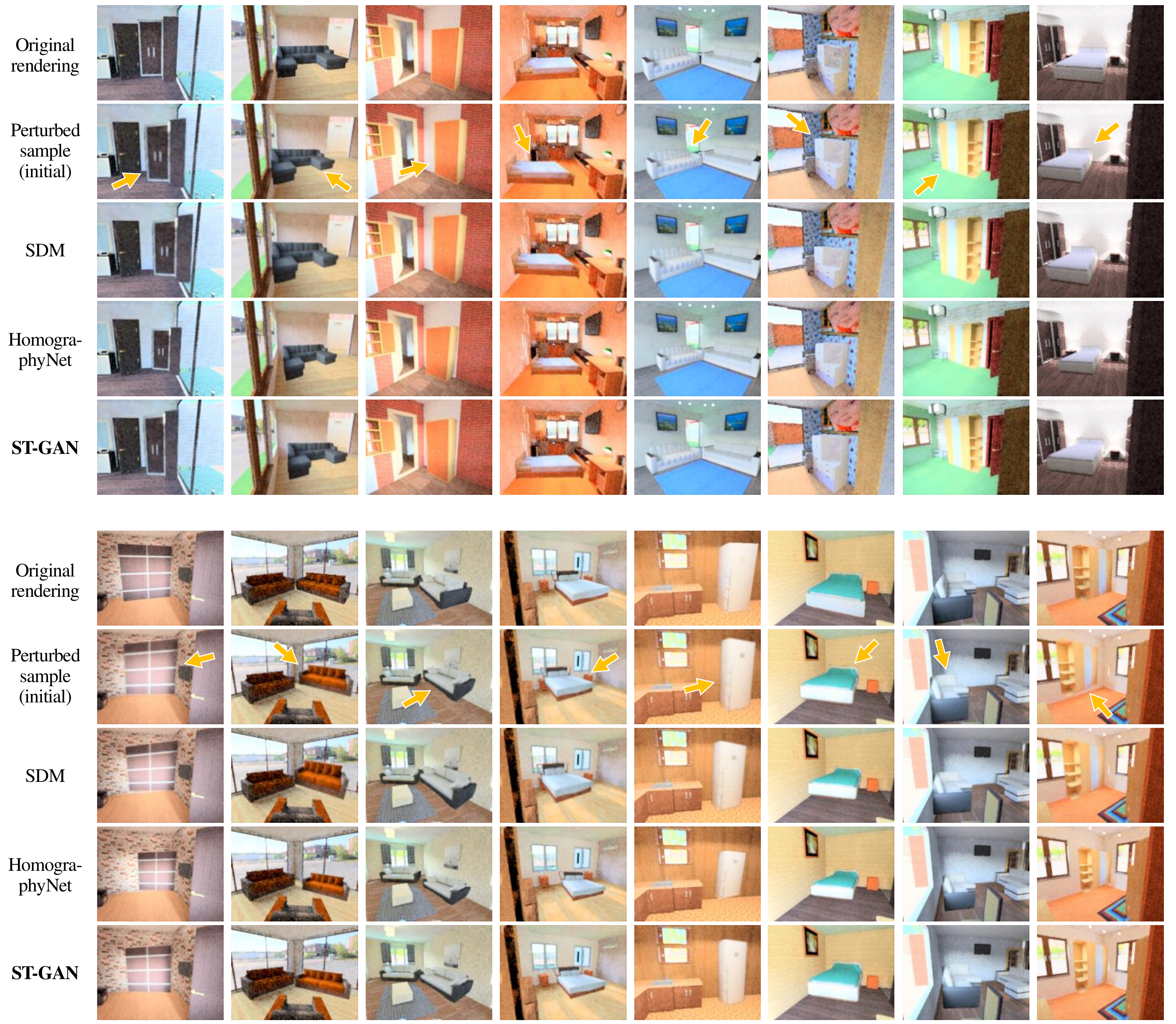}
\caption{Additional qualitative results from the indoor object experiment (test set).
The yellow arrows in the second row point to the composited foreground objects.}
\label{fig:suncg_results_supp}
\end{figure*}

We include additional qualitative results from the indoor object experiment in Fig.~\ref{fig:suncg_results_supp}.
Compared to the baselines, ST-GAN consistently predicts more realistic geometric corrections in most cases.

\subsection*{A.5. Additional Glasses Results}

\begin{figure*}[t!] \center
\includegraphics[width=\linewidth]{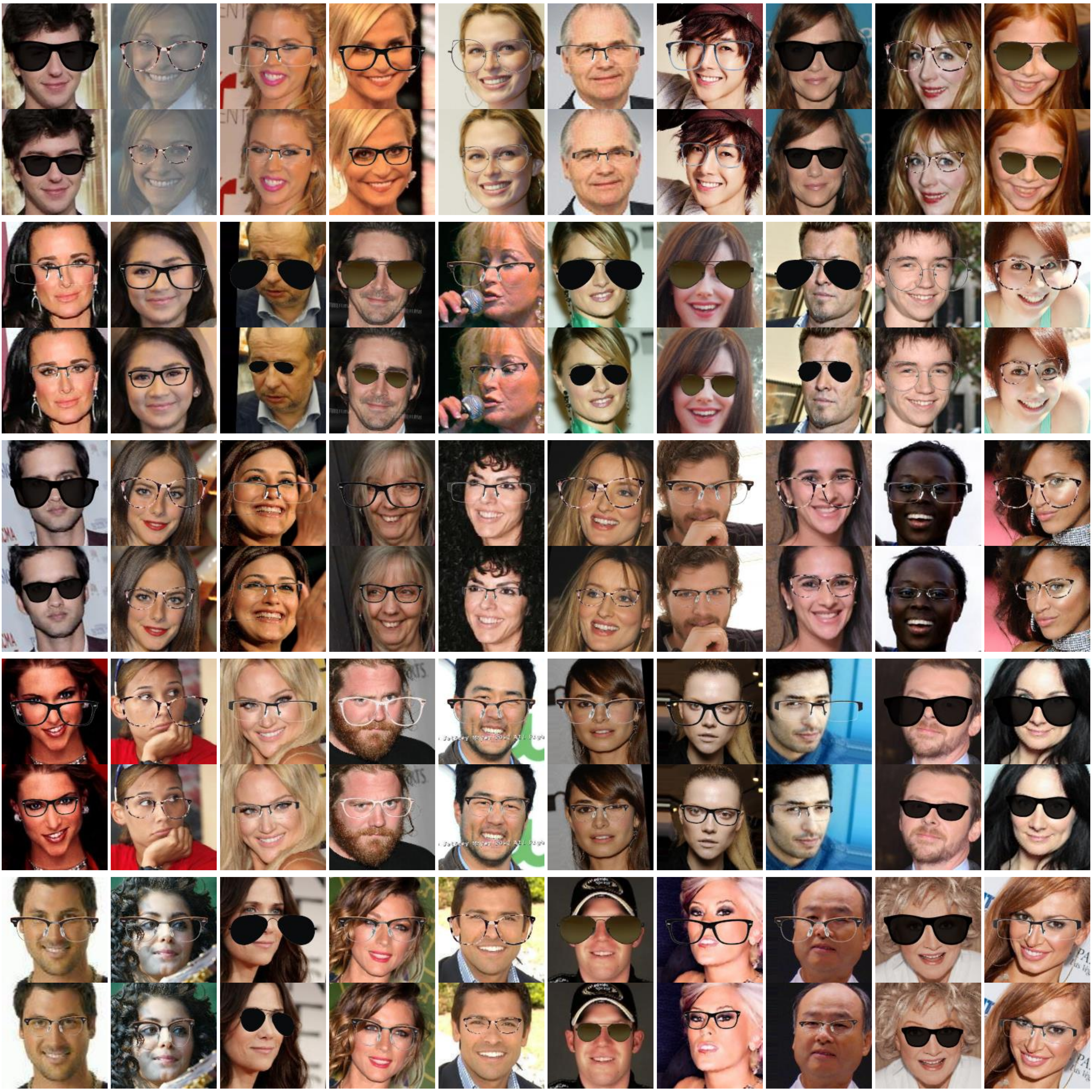}
\caption{Additional qualitative results from the glasses experiment (test set).
The top row indicates the initial composite, and the bottom row indicates the ST-GAN output.}
\label{fig:glasses_results_supp}
\end{figure*}

We also include additional qualitative results from the glasses experiment in Fig.~\ref{fig:glasses_results_supp}.
We re-emphasize that the training data here is unpaired and there is no information in the dataset about where the glasses are placed.
Despite these, ST-GAN is able to consistently match the initial glasses foreground to the background faces.


\end{document}